 \documentclass[final,5p,times,twocolumn,authoryear]{elsarticle}

\usepackage{amsmath,amssymb,amsfonts}

\usepackage{natbib}
\setcitestyle{numbers}
\setcitestyle{square}

\usepackage{booktabs}
\usepackage{graphicx}
\usepackage[none]{hyphenat}
\usepackage{multirow}
\usepackage{subcaption}

\journal{Computer Methods and Programs in Biomedicine}

\def\BibTeX{{\rm B\kern-.05em{\sc i\kern-.025em b}\kern-.08em
		T\kern-.1667em\lower.7ex\hbox{E}\kern-.125emX}}
\begin{document}
	
\begin{frontmatter}

\title{Window-Based Feature Engineering for Cognitive Workload Detection}

\author[first]{Andrew Hallam}

\author[first]{R G Gayathri}%

\author[first]{Glory Lee}%

\author[first]{Atul Sajjanhar\fnref{label1}}
\affiliation[first]{organization={School of Information Technology, },
            addressline={Deakin University}, 
            city={Geelong},
            state={VIC},
            country={Australia}}
\fntext[label1]{corresponding author at School of Information Technology, Deakin University, Geelong, Victoria, Australia, atul.sajjanhar@deakin.edu.au%
}

\begin{abstract}
	Cognitive workload is a topic of increasing interest across various fields such as health, psychology, and defense applications. In this research, we focus on classifying cognitive workload using the COLET dataset, employing a window-based approach for feature generation and machine/deep learning techniques for classification. We apply window-based temporal partitioning to enhance features used in existing research, followed by machine learning and deep learning models to classify different levels of cognitive workload. The results demonstrate that deep learning models, particularly tabular architectures, outperformed traditional machine learning methods in precision, F1-score, accuracy, and classification precision. This study highlights the effectiveness of window-based temporal feature extraction and the potential of deep learning techniques for real-time cognitive workload assessment in complex and dynamic tasks.
\end{abstract}	
		
\end{frontmatter}
	
	\section{Introduction}
	Cognitive workload (CW) classification is an emerging research topic that encapsulates a broad spectrum of fields, including healthcare, psychology, and ergonomics\cite{torkamani2022methods}\cite{joseph2020potential}. 
    With advancements in deep learning, unsynchronized raw eye movement data from real-world datasets can possibly be handled and analyzed. Nonetheless, the inherent complexity and heterogeneity of raw eye movement data comprising various types like positional and temporal data necessitate intricate patterns that typically require extensive pre-processing or sophisticated feature engineering \cite{murthy2022deep}. 
	
	
	Previous research has explored CW classification using physiological signals such as EEG, heart rate, and eye-tracking data \cite{Zhou_eeg} \cite{Stikic_heart_rate}. 
    Eye-tracking metrics, such as pupil diameter and fixation patterns, have been shown to correlate strongly with CW levels across various tasks. Traditional machine learning approaches \cite{shojaeizadeh2019detecting} \cite{ramakrishnan2021cognitive} \cite{liu2022assessing}, such as Support Vector Machines (SVM), Random Forests (RF), and Naive Bayes (NB), have been applied to classify cognitive workload using features like pupil dilation and fixation duration \cite{bozkir2019person}\cite{rizzo2022machine}\cite{ktistakis2022colet}. Deep learning models have recently been used to capture complex temporal patterns in eye-tracking data. 
	
	Temporal segmentation divides eye movement data into non-overlapping, discrete length time windows. It enables a focused analysis of eye-tracking metrics across specific periods, reviewing how cognitive load fluctuates over time. By segmenting data into manageable windows, researchers can observe trends or detect spikes in CW, especially during critical moments of task performance. 
    In this research, we explore eye-tracking technology and time-series feature generation methods to develop an effective CW classification system. Specifically, we investigate tabular deep learning compared to traditional machine learning models. Our investigation also focuses on applying tabular deep learning architecture, which has demonstrated promising results, outperforming existing deep learning models and traditional machine learning models across multiple existing tabular datasets \cite{arik2021tabnet}. We propose a window-based approach for feature generation and apply machine learning and deep learning techniques for CW classification. Our contributions include: 
	
	\begin{itemize}
    
		\item a novel method for segmenting continuous eye-tracking data for exploiting the temporal correlation of eye-tracking data that had not been fully explored in CW detection.
		
	      \item an effective feature set generation technique for tabular-based deep learning and performance analysis compared to traditional machine learning models.

        \item an extensive comparative study of both traditional machine learning models and advanced deep learning models to evaluate their effectiveness in predicting CW using the COLET temporal features. 

	\end{itemize}
	The remainder of this paper is structured as follows. Section \ref{sec_Lit_Review} summarizes the key concepts adopted in this work. Section \ref{sec_methodology} explains the proposed method, which includes the window-based feature engineering for CW detection. Section \ref{sec_exp_setup} evaluates performance using various metrics. Section \ref{sec_conclusion} concludes the study.
	
\section{Background}\label{sec_Lit_Review}
    In this section, we provide an overview of the key features used in the eye movement tracking dataset, traditional machine learning approaches, and state-of-the-art tabular deep learning approaches in the context of CW classification. 
	
	
	 \subsection{Structured Eye-tracking}
   
	Ktistakis et al. \cite{ktistakis2022colet} established the COgnitive workLoad estimation based on Eye-Tracking (COLET) dataset in this field. The dataset includes synchronized eye tracking and task performance data, which allows researchers to investigate how the cognitive workloads of various users are classified. The collected eye and gaze movement recordings of subjects, along with their performance scores when solving CAPTCHA-like puzzles related to visual search workload conditions, were categorized into three levels: low (L), medium (M), and high (H). Performance metrics such as errors, reaction time, and inverse efficiency score, alongside physical measures including fixations, saccades, blinks, and pupil size, are recorded to assess workload levels. These measures are summarized in Table \ref{tab:colet_measures}.


	
	
	\begin{table}[!hbtp]
		\resizebox{\columnwidth}{!}{%
             \begin{tabular} {@{}lp{4.2cm}l@{}} 
				\toprule
				\textbf{Key Measure} & \textbf{Definition}                                     & \textbf{Aggregate}                                                         \\ \midrule
				Fixation & Duration of focus on a single point. & \begin{tabular}[c]{@{}l@{}}Frequency, Duration\end{tabular} \\ \hline
				Saccade              & Rapid eye movements between fixations.                  & \begin{tabular}[c]{@{}l@{}}Frequency, Duration, Velocity\end{tabular} \\ \hline
				Blink                & Frequency of blinking during task performance.      & \begin{tabular}[c]{@{}l@{}}Frequency, Duration\end{tabular}             \\ \hline
				Pupil                & Pupil size changes in response to cognitive demands. & Diameter                                                                   \\ \bottomrule 
			\end{tabular}%
		}
		\caption{Key cognitive workload measures and aggregates from COLET}
		\label{tab:colet_measures}
	\end{table}

\subsection{Tabular Deep Learning}

The literature shows that machine learning approaches are strong predictive models for CW. Table \ref{tab:summary_lit3} summarizes existing model performances, with an emphasis on evaluation measures in accuracy and F1 scores. While traditional machine learning has been widely used for CW detection, deep learning models remain underexplored in this domain. Deep learning methodology has shown strong performance with large structured datasets with robust features \cite{Sun2017datasize}, making it a promising alternative for workload classification. However, most deep learning studies on eye tracking focus on image-based models \cite{Saxena_image_eye_data} and have not been applied to the domain of CW detection \cite{rizzo2022machine}. To bridge this gap, we investigate the potential of tabular deep learning for CW detection, inspired by the promising performance in prior studies that applied deep learning methodology to tabular datasets \cite{Saravanan_cnn}. 

Our research aims to generate more effective features for deep learning-based CW detection by fully leveraging COLET's tabular eye movement data, particularly to address its post-aggregation limitations. By incorporating temporal structure, we enhance feature extraction, making the dataset more suitable for deep learning models. In this work, we focus on TabNet \cite{arik2021tabnet}, a deep learning architecture designed for tabular data. TabNet integrates deep neural networks, a gradient descent optimization method, and sequential attention mechanisms. Its key components include a feature transformer capturing patterns within the data through sequential attention, a decision point that produces masked and sparsity-masked features following feature transformations, and an attentive transformer that applies attention mechanisms to weigh feature importance. Its aggregator and predictor combine attention-weighted outputs to generate predictions through fully connected layers. We aim to enhance deep learning effectiveness in CW classification using TabNet’s ability to selectively focus on relevant features, providing an alternative to traditional machine learning approaches.
    
\section{Proposed Method}\label{sec_methodology}
\begin{figure}[!thbp]
		\centering
		\includegraphics[scale=0.28]{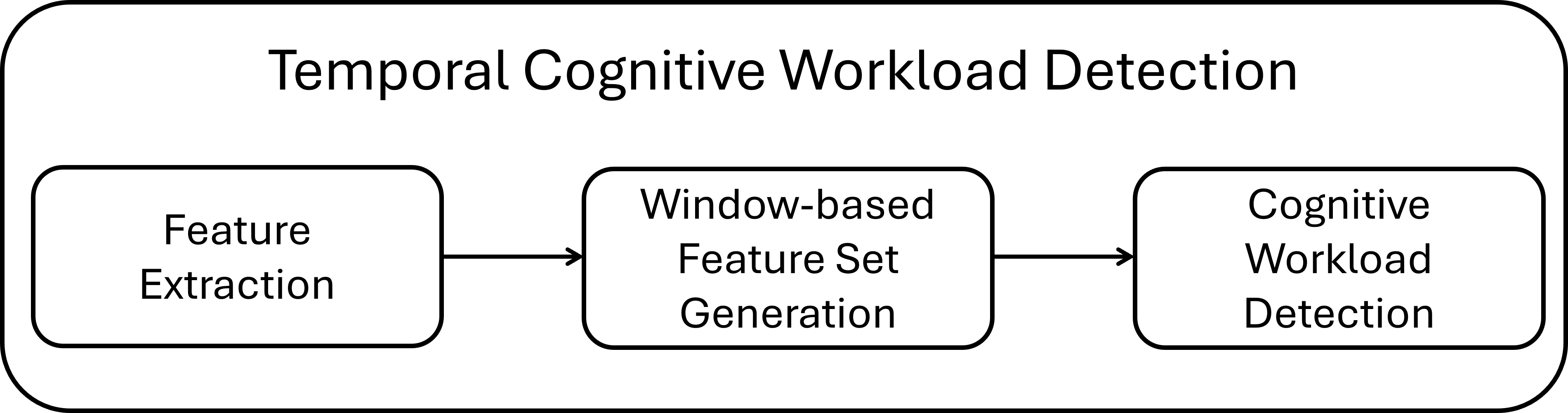}
		\caption{Overview of the proposed method}
		\label{fig:overall-approach}
	\end{figure}

In this study, we propose a CW classification system using eye-tracking data and time-series window-based feature generation methods. The COLET dataset\cite{ktistakis2022colet}, which contains eye-tracking data collected under real-world driving conditions, is the foundation for this research. The proposed method starts with the feature engineering of the COLET data, followed by the CW detection. Figure \ref{fig:overall-approach} shows the overview of the method followed in this work.

    Each step is explained in detail in the following sections. The COLET dataset is used to extract a set of features from the original work. The first step involves extracting relevant features from the eye-tracking data, which is explained in Section \ref{ftr_sec_gen}. This step is followed by transforming the feature set to a time-series feature set using a window-based approach; the details are given in Section \ref{sec_wndw_ftr_gen}. Section \ref{sec_dl_cw_detection} explains the CW detection using deep-learning models.
	
\subsection{Feature Extraction}\label{ftr_sec_gen}
\begin{figure}[tbph!]
		\centering
		\includegraphics[scale=0.29]{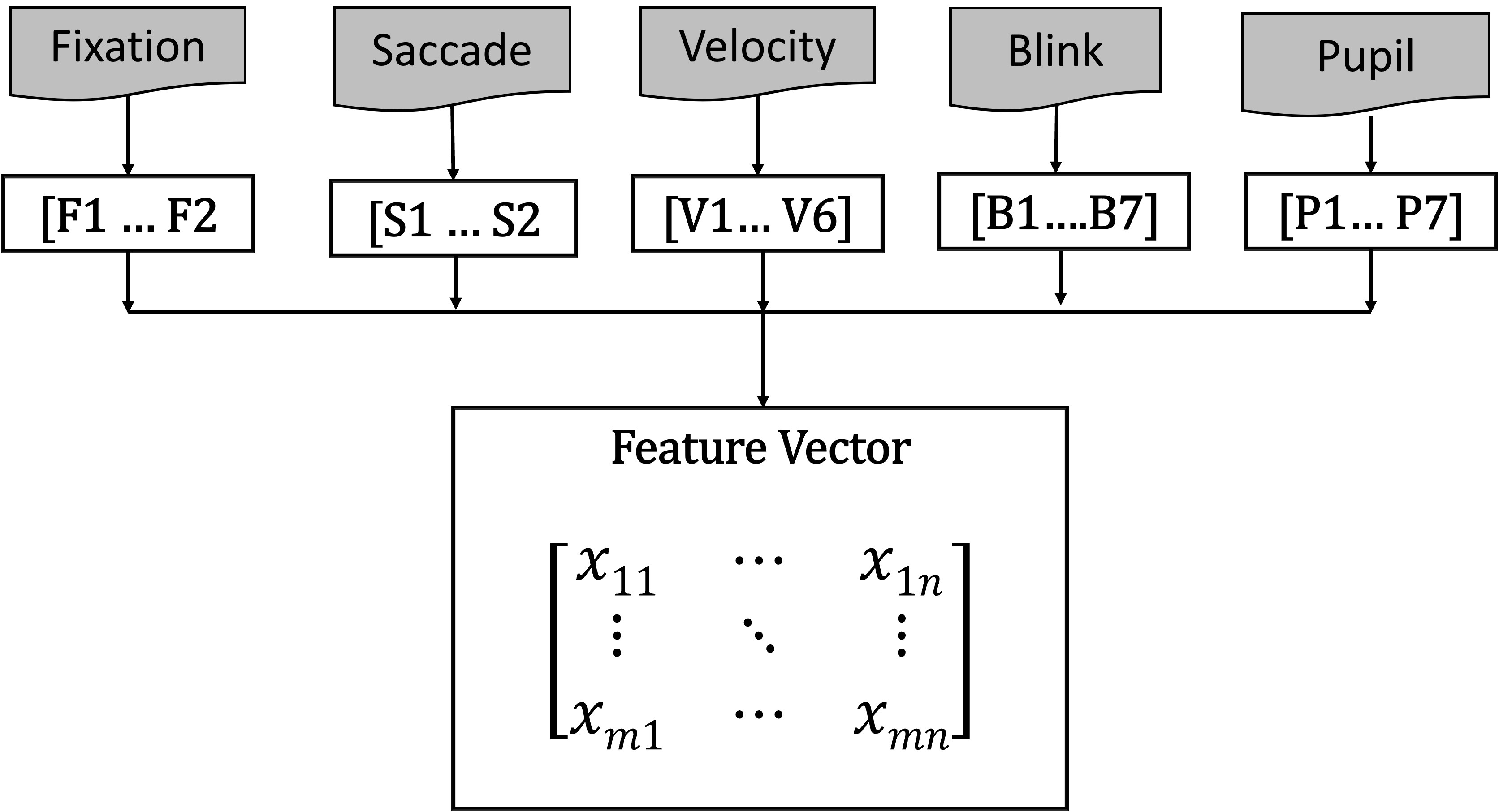}
		\caption{Feature extraction from COLET Data}
		\label{fig:featureset}
	\end{figure}

	Identifying appropriate features linked to the target variable is essential for building machine learning models. The raw eye-tracking data from the COLET dataset is cleaned by handling missing values, outliers, and noisy data. The key measures mentioned in the Table \ref{tab:colet_measures} from the COLET dataset are used to extract the features. Comprehensive sets of features are extracted from the COLET data as shown in Figure \ref{fig:featureset}. Details of the features from each key measure are consolidated in Table \ref{tab:windw_ftr_set}.
	
    \begin{table*}[!hbtp]
    \centering
    \small
    \begin{tabular}{lll} 
    \toprule
    \textbf{Criteria} & \textbf{Feature\_ID} & \textbf{Description} \\ 
    \midrule
    Fixation & F1 & \begin{tabular}[c]{@{}l@{}}Number of fixations identified \\ (slow eye movements)\end{tabular} \\
     & F2 & Number of fixations identified per second (slow eye movements) \\
    Saccade & S1 & number of saccades identified per second (fast eye movements) \\
     & S2 & Number of saccades identified (fast eye movements) \\
    Velocity & V1 & Lowest eye movement velocity \\
     & V2 & Fastest eye movement velocity \\
     & V3 & skewness of maximum eye velocity \\
     & V4 & kurtosis of maximum eye velocity \\
     & V5 & skewness of minimum eye velocity \\
     & V6 & kurtosis of minimum eye velocity \\
    Pupil diameter & P1 & average pupil diameter identified \\
     & P2 & maximum pupil diameter identified \\
     & P3 & minimum pupil diameter identified \\
     & P4 & sknewness of the minimum pupil diameter identified \\
     & P5 & kurtosis of the minimum pupil diameter identified \\
     & P6 & skewness of the maximum pupil diameter identified \\
     & P7 & kurtosis of the maximum pupil diameter identified \\
    \multirow{7}{*}{Blink duration} & B1 & average blink duration \\
     & B2 & maximum blink duration \\
     & B3 & minimum blink duration \\
     & B4 & skewness of minimum blink duration \\
     & B5 & kurtosis of minimum blink duration \\
     & B6 & skewness of maximum blink duration \\
     & B7 & kurtosis of maximum blink duration \\
    Cognitive workload & Target & measure of task cognitive workload, numeric \\
    \bottomrule
    \end{tabular}
	\caption{Window-based feature set}
	\label{tab:windw_ftr_set}
	\end{table*}

	\subsection{Window-based Feature Engineering} \label{sec_wndw_ftr_gen}
	
	Eye-tracking data collection involves capturing various time-sensitive data, such as pupil dilation, fixation duration, saccade amplitude, blink rate, and gaze patterns. These temporal metrics provide crucial insights for deep learning models into the user's visual processing and cognitive states. Since eye-tracking data are typically a continuous stream, a window-based approach is essential to partition the data into manageable, interpretable intervals for analysis. 

	In this work, we propose a window-based approach for eye movement tracking in deep learning-based CW detection. This approach involves partitioning the eye-tracking data into fixed-size, non-overlapping windows and analyzing the features within each window to detect changes in cognitive load. This method allows capturing the temporal dynamics of eye movements over time, which can be crucial for identifying patterns associated with varying levels of CW. This window-based approach allows deep learning models to effectively capture the temporal patterns in eye movements, providing a robust framework for CW detection. 
	
	In window-based segmentation, the continuous stream of eye-tracking data is divided into fixed-length windows, such as 1-second or 2-second intervals. Temporal patterns and workload fluctuations can be captured more effectively by segmenting the data into these windows, allowing for a detailed analysis of how eye movements evolve. This segmentation is crucial for enabling deep learning models to process the data and detect shifts in cognitive effort with greater accuracy. 
	
	In addition to extracting features from each window, the temporal evolution of features across windows can be captured (e.g., rate of change of pupil dilation over consecutive windows). Features like mean, standard deviation, and skewness across each window help capture workload variations. For each window, we extract statistical and temporal features such as: 
	\begin{itemize}
		\item \textbf{Pupil Dilation Features:} Mean, standard deviation, and skewness of pupil diameter. 
		
		\item \textbf{Fixation Features:} Mean fixation duration, fixation count, and velocity. 
		
		\item \textbf{Saccade Features:} Mean saccade amplitude, velocity, and duration. 
		
		\item \textbf{Blink:} Mean, minimum, maximum, standard deviation, and skewness of blink duration 
	\end{itemize}
	
   For instance, longer fixation durations or larger pupil dilations in particular windows can signal moments of increased task complexity or mental effort. Examining these changes within each time window makes associating cognitive load spikes with specific task elements or challenges possible. Additionally, this window-based segmentation enables real-time CW detection, which is particularly valuable for adaptive systems. These systems can dynamically adjust task difficulty or provide real-time assistance based on workload variations detected in each window. 
	
    \subsection*{Key Benefits of Adopting the Window-Based Method}
	The window-based approach offers several advantages for CW detection. One key benefit is its ability to preserve temporal dynamics by segmenting data into windows, allowing it to capture patterns over time that are critical for detecting shifts in cognitive load. Additionally, the approach is highly adaptable, as the window size can be adjusted based on the complexity of the task, enabling the model to account for both short-term and long-term workload variations. Moreover, this method provides flexibility in model selection, allowing researchers to apply different deep-learning models depending on the type of eye movement data and the nature of the task. This adaptability makes the window-based approach versatile for various CW detection scenarios. The proposed method of maintaining this temporal aspect as a feature is to introduce a window-based approach. 
    
	
	\subsection*{Window Generation}  
	The first (F\_Time) and last timestamp (L\_Time) range is identified and grouped by participant and task. This will give us the range for each participant’s task, for example, participant 1, 	task 1, participant 1, task 2, and so on. The range is divided by $n$ for an even number of $n$ windows. 
	
	\textbf{First and Last Timestamp: }Each window represents a segment of time over the task duration. For every participant, identify the first timestamp (F\_Time), which is the earliest time they engaged in the task, and the last timestamp (L\_Time), which is the latest time they completed any action in the task.
	
	\[
	F\_Time = \min(\text{timestamps for all actions by participant})
	\]
	\[
	L\_Time = \max(\text{timestamps for all actions by participant})
	\]
	

	\textbf{Calculate the Range: }Once you have the first and last timestamps, calculate the total duration of the task for that participant by subtracting the last timestamp from the first timestamp as given in Equation \ref{eq_1}:
	\begin{equation} 
		\label{eq_1}
		Range = L\_Time - F\_Time
	\end{equation}

	\textbf{Divide the Range into $X$ Windows: }After calculating the total range, divide it into $X$ equal-time windows where $X$ is the number of desired intervals. This gives you time segments to analyze the participant’s actions during the task. Each window will have a duration of $W$ as given in Equation \ref{eq_2}:
	
	\begin{equation}
		\label{eq_2}
		W = \frac{\text{Range}}{X}
	\end{equation}
	
	These windows help evaluate patterns, behaviors, or progress over time, breaking down the participant’s task engagement into smaller, more manageable time intervals. This approach helps track participant activity over a defined time by dividing their engagement into equally spaced intervals, and can offer insights into how their performance evolves, which helps better understand task execution patterns.
	
	\begin{figure}[!h]
		\centering
		\includegraphics[width=0.8\linewidth]{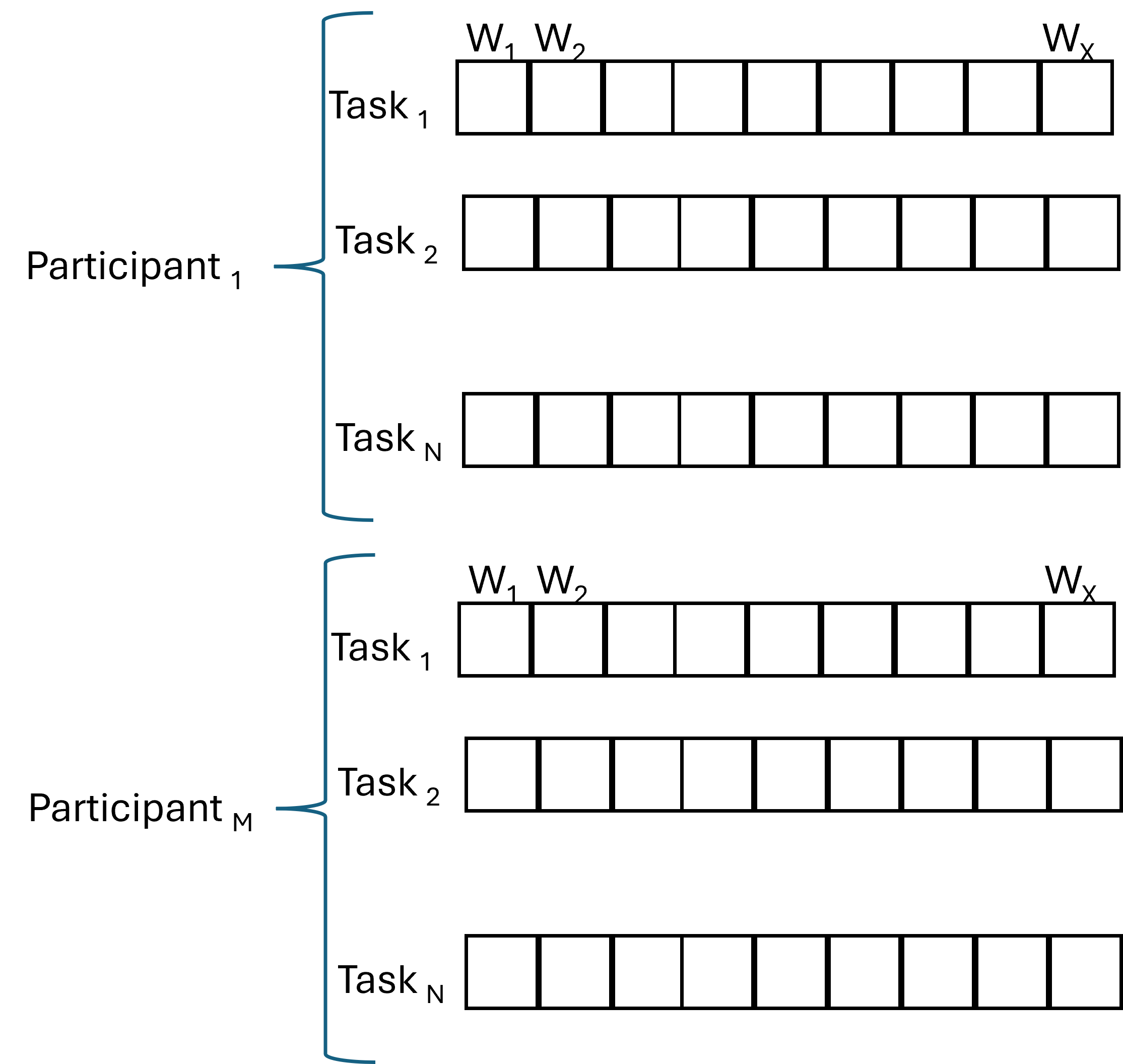}
		\caption{Window Generation}
		\label{fig:window}
	\end{figure}
	
	A value of $1$ to $n$ is applied to these windows, grouped as noted above. This means that the first window in the first task, for the first participant, is given a value of 1, and the last is given a value of 16. The same applies across all participants and tasks; for example, participant 2, task 3, will have the first window of 1 and the last window of 16. This allows for easier joins and provides comparable information when comparing across windows, as the first window will be the earliest for each task and participant, and the last window will be the latest.

	\subsection{Cognitive Workload Detection} \label{sec_dl_cw_detection}
	
	Traditional machine learning techniques, such as gradient-boosted decision trees (GBDT), have dominated tabular data modeling over the past decade, often outperforming deep learning methods. Recently, several studies have explored deep learning networks for tabular data, with some claiming to surpass the performance of GBDT. These works suggest that regularized deep learning models can effectively handle tabular data. 
	
	
	In this work, we adopt four different architectures to use deep learning (DL) models for CW detection: Multi-layer Perceptron (MLP), Self-Normalizing Neural Network (SNN), 1D Convolutional Neural Network (1DCNN), and TabNet. The selection of each model is based on its distinct ability to handle high-dimensional, complicated data structures, which are necessary for precise CW detection.

\section{Experiment and Result Analysis}\label{sec_exp_setup}
	In this section, we evaluate the performance of the proposed method. We describe the dataset, followed by the models that are used for training and performance evaluation. We have implemented the method using Python programming language, Tensorflow, and Keras.  
	
	\subsection{Training and Testing Conditions}
	
	Very few appropriate, tabular datasets are available in the domain of CW detection. The majority of identified datasets are image datasets. One recent and popular tabular dataset is the COLET\cite{ktistakis2022colet} dataset.  
	
	\subsection{Performance Evaluation}
    The MLA dataset refers to the work of Rizzo et al. \cite{rizzo2022machine}, while the private datasets correspond to research by Bozkir, Geisler, and Kasneci \cite{bozkir2019person} and Joseph and Murugesh \cite{joseph2020potential}, respectively.  Table \ref{tab:summary_lit3} summarizes findings from prior studies. 

    \begin{table}[!hbtp]
		\centering
        \small
		\begin{tabular}{@{}llll@{}}
			\toprule
			\textbf{Model} & \textbf{Accuracy} & \textbf{F1}     & \textbf{Dataset} \\ \midrule
			DT             & 73.4\% (H,   L)   & 71.65\% (H, L)  & Private         \\
			& 86.8\% (H,   L)   & Not reported    & Private          \\
			& 74\% (L/M, H)     & 73\% (L/M,   H) & COLET            \\ \hline
			SVM            & 93.3\% (L/M,   H) & Not reported    & Private          \\
			& 80.7\% (H,L)      & 80.98\% (H,L)   & Private          \\
			& 67\% (R, RWI)     & 68\% (R, RWI)   & MLA            \\
			& 72\% (N, NWI)     & 68\% (R, RWI)   & MLA              \\
			& 69\% (L,M)        & 69\% (L,M)      & COLET            \\ \hline
			GNB            & 59\% (L,M/H)      & 59\% (L,M/H)    & COLET            \\
			& 88\% (L,H)        & 86\% (L,H)      & COLET            \\ \hline
			LR             & 51\% (L,M,H)      & 50\% (L,M,H)    & COLET            \\
			& 68\% (N, NWI)     & 69\% (N, NWI)   & MLA              \\
			& 68\% (R,   RWI)   & 68\% (R, RWI)   & MLA               \\ \hline
			Ensemble       & 58\% (L,H)        & 57\% (L,H)      & COLET            \\ \hline
			RF             & 84\% (L,H)        & 84\% (L,H)      & COLET            \\
			& 62\% (R, RWI)     & 67\% (R, RWI)   & MLA              \\
			& 62\% (N,   NWI)   & 59\% (N, NWI)   & MLA              \\ \bottomrule
		\end{tabular}
		\caption{Summary of model performance in the existing literature.}
		\label{tab:summary_lit3}
	\end{table}

    A comprehensive set of experiments was conducted on the COLET data on the original features proposed by the authors, as well as the window-based feature engineering proposed in this work. Section \ref{benchmark_comparison} provides the comparison of the original paper with the features that the authors have reproduced. The comparison is done using the binary and the multi-class classification. The window-based feature set is experimented using the binary and the multi-class classification on the ML and DL models, and the results are summarized in Section \ref{window_multiclass}. We use text highlighted in bold to depict the interesting results in the tables with experimental results. 
	
	\subsection{Benchmark Evaluation for Binary Classification}\label{benchmark_comparison}
	
	In the original work\cite{ktistakis2022colet}, the authors trained and tested eight classifiers. In this work, we used the GNB, SVM, k-NN, NB, and LR classifiers for comparison. The same hyperparameters from the \cite{ktistakis2022colet} are used to reproduce the results. 
    
    The main difference between the COLET experiments \cite{ktistakis2022colet} and the proposed methodology is the window-based feature sets and the proposed CW detection using tabular DL models, including baseline deep learning models, self-normalizing neural network, 1DNN, and multilayer perceptron, and the proposed TabNet model. The specific transformations applied to the COLET dataset in the original work remain unclear, as the model dataset is unavailable, and only the raw data is accessible. Consequently, slight differences are expected when comparing the conventional approach with the COLET authors' method.
	
	The multiclass classifiers predicted the target labels as follows. For COLET, this was low (1), medium (2), and high (3) CW. For the binary classifiers, classes 1 and 2 are merged, and class 3 remains as the second binary condition. This matches the methodology of the COLET authors \cite{ktistakis2022colet}. The class conditions tested are as follows: 
	\begin{itemize}
		\item C1/C2: Classification between classes C1 and C2 - Binary
		\item C1/C3: Classification between classes C1 and C3 - Binary
		\item C2/C3: Classification between classes C2 and C3 - Binary
		\item C1,2/C3: Classification between combined classes C1 and C2 versus class C3 - Binary
        \item C1/C2/C3: Classification between classes C1 versus C2 versus class C3 - Multiclass\end{itemize}

\subsubsection{Binary classification using ML models}\label{Performance of classification using the ML}
	Here, we study the performance of the ML models for binary classification. By comparing results to the COLET author's results, we can understand how reasonable the feature extraction has been. When removing the temporal aspect (windows), the results largely mimic those of the COLET authors, with some exceptions. Given the limitations noted in the section, including non-availability of transformed data, missing data, and interpretation of saccade and fixation threshold identification, it is difficult to entirely replicate the data and transformations completed by the authors, but as shown in Table \ref{tab:compare-benchmark_ML}, the results presented here closely align with the results of the COLET authors. It is, therefore, reasonable to conclude that the resulting dataset is similar.  

\begin{table}[!thbp]
\resizebox{\columnwidth}{!}{%
\begin{tabular}{|c|l|cccc|c|}
\hline
\multirow{2}{*}{\textbf{Model}} &
  \multicolumn{1}{c|}{\multirow{2}{*}{\textbf{Metric}}} &
  \multicolumn{4}{c|}{\textbf{Binary}} &
  \multicolumn{1}{l|}{\textbf{Multi-class}} \\ \cline{3-7} 
 &
  \multicolumn{1}{c|}{} &
  \multicolumn{1}{c|}{\textbf{C1/C2}} &
  \multicolumn{1}{c|}{\textbf{C1/C3}} &
  \multicolumn{1}{c|}{\textbf{C2/C3}} &
  \textbf{C1,2/C3} &
  \textbf{C1/C2/C3} \\ \hline
\multirow{4}{*}{\textbf{SVM}}  & Precision & \multicolumn{1}{c|}{0.68} & \multicolumn{1}{c|}{0.93} & \multicolumn{1}{c|}{0.63} & 0.66 & 0.55 \\ 
                               & Recall~   & \multicolumn{1}{c|}{0.7}  & \multicolumn{1}{c|}{0.97} & \multicolumn{1}{c|}{0.62} & 0.63 & 0.56 \\ 
                               & F1        & \multicolumn{1}{c|}{0.68} & \multicolumn{1}{c|}{0.95} & \multicolumn{1}{c|}{0.63} & 0.64 & 0.53 \\  
                               & Accuracy  & \multicolumn{1}{c|}{0.7}  & \multicolumn{1}{c|}{0.96} & \multicolumn{1}{c|}{0.67} & 0.77 & 0.58 \\ \hline
\multirow{4}{*}{\textbf{CART}} & Precision & \multicolumn{1}{c|}{0.55} & \multicolumn{1}{c|}{0.87} & \multicolumn{1}{c|}{0.61} & 0.62 & 0.63 \\ 
                               & Recall~   & \multicolumn{1}{c|}{0.55} & \multicolumn{1}{c|}{0.83} & \multicolumn{1}{c|}{0.62} & 0.63 & 0.55 \\ 
                               & F1        & \multicolumn{1}{c|}{0.54} & \multicolumn{1}{c|}{0.85} & \multicolumn{1}{c|}{0.61} & 0.63 & 0.56 \\ 
                               & Accuracy  & \multicolumn{1}{c|}{0.57} & \multicolumn{1}{c|}{0.89} & \multicolumn{1}{c|}{0.63} & 0.72 & 0.57 \\ \hline
\multirow{4}{*}{\textbf{GNB}}  & Precision & \multicolumn{1}{c|}{0.77} & \multicolumn{1}{c|}{0.93} & \multicolumn{1}{c|}{0.44} & 0.69 & 0.76 \\ 
                               & Recall~   & \multicolumn{1}{c|}{0.8}  & \multicolumn{1}{c|}{0.98} & \multicolumn{1}{c|}{0.46} & 0.66 & 0.62 \\ 
                               & F1        & \multicolumn{1}{c|}{0.76} & \multicolumn{1}{c|}{0.96} & \multicolumn{1}{c|}{0.44} & 0.66 & 0.61 \\ 
                               & Accuracy  & \multicolumn{1}{c|}{0.77} & \multicolumn{1}{c|}{0.96} & \multicolumn{1}{c|}{0.53} & 0.77 & 0.66 \\ \hline
\multirow{4}{*}{\textbf{KNN}}  & Precision & \multicolumn{1}{c|}{0.86} & \multicolumn{1}{c|}{1}    & \multicolumn{1}{c|}{0.88} & 0.93 & 0.73 \\  
                               & Recall~   & \multicolumn{1}{c|}{0.9}  & \multicolumn{1}{c|}{1}    & \multicolumn{1}{c|}{0.72} & 0.72 & 0.71 \\  
                               & F1        & \multicolumn{1}{c|}{0.89} & \multicolumn{1}{c|}{1}    & \multicolumn{1}{c|}{0.74} & 0.77 & 0.72 \\  
                               & Accuracy  & \multicolumn{1}{c|}{0.87} & \multicolumn{1}{c|}{1}    & \multicolumn{1}{c|}{0.77} & 0.87 & 0.72 \\ \hline
\multirow{4}{*}{\textbf{LR}}   & Precision & \multicolumn{1}{c|}{n/a}  & \multicolumn{1}{c|}{n/a}  & \multicolumn{1}{c|}{n/a}  & n/a  & 0.55 \\  
                               & Recall~   & \multicolumn{1}{c|}{n/a}  & \multicolumn{1}{c|}{n/a}  & \multicolumn{1}{c|}{n/a}  & n/a  & 0.56 \\ 
                               & F1        & \multicolumn{1}{c|}{n/a}  & \multicolumn{1}{c|}{n/a}  & \multicolumn{1}{c|}{n/a}  & n/a  & 0.53 \\
                               & Accuracy  & \multicolumn{1}{c|}{n/a}  & \multicolumn{1}{c|}{n/a}  & \multicolumn{1}{c|}{n/a}  & n/a  & 0.57 \\ \hline
\end{tabular}%
}
\caption{Binary classification using the ML models on benchnamrk dataset}
\label{tab:compare-benchmark_ML}
\end{table}

Table \ref{tab:compare-benchmark_ML} shows the performance metrics for five machine learning models: SVM, CART, GNB, KNN and LR. Four binary classification conditions are used to evaluate these models: "C1/C2," "C1/C3," "C2/C3," and "C1,2/C3." Each condition represents a distinct categorization scenario with various class combinations. The C1/C3 condition yields the highest precision, recall, F1, and accuracy values for SVM. The C1/C3 condition is where CART performs best, with an accuracy of 0.89. Like SVM, GNB equally performs best in the C1/C3 condition, with an accuracy of 0.96. It has relatively poor metrics and struggles in the C2/C3 condition. KNN performs exceptionally well in this particular classification scenario, achieving perfect scores (1.00) for precision, recall, F1, and accuracy in the C1/C3 condition.


    \subsubsection{Binary Classification on DL models}\label{Performance of classification using the ML} 
	
    Table \ref{tab:compare-benchmark_DL} compares the performance of four deep learning models (TabNet, SNN, MLP, and 1DCNN) across various class conditions (C1/C2, C1/C3, C2/C3, C1,2/C3) using four metrics: Precision, Recall, F1 Score, and Accuracy. With its best accuracy (0.75) in the C1/C3 condition, TabNet performs rather poorly overall, particularly in the C1/C2 and C1,C2/C3 conditions. 
\begin{table}[!hbtp]
\resizebox{\columnwidth}{!}{%
\begin{tabular}{|c|l|cccc|c|}
\hline
\multirow{2}{*}{\textbf{Model}} &
  \multicolumn{1}{c|}{\multirow{2}{*}{\textbf{Metric}}} &
  \multicolumn{4}{c|}{\textbf{Binary}} &
  \multicolumn{1}{l|}{\textbf{Multi-class}} \\ \cline{3-7} 
 &
  \multicolumn{1}{c|}{} &
  \multicolumn{1}{c|}{\textbf{C1/C2}} &
  \multicolumn{1}{c|}{\textbf{C1/C3}} &
  \multicolumn{1}{c|}{\textbf{C2/C3}} &
  \textbf{C1,2/C3} &
  \textbf{C1/C2/C3} \\ \hline
\multirow{4}{*}{\textbf{TabNet}} & Precision & \multicolumn{1}{c|}{0.69} & \multicolumn{1}{c|}{0.66} & \multicolumn{1}{c|}{0.64} & 0.55 & 0.28 \\ 
                                 & Recall~   & \multicolumn{1}{c|}{0.58} & \multicolumn{1}{c|}{0.64} & \multicolumn{1}{c|}{0.58} & 0.56 & 0.36 \\ 
                                 & F1-Score  & \multicolumn{1}{c|}{0.4}  & \multicolumn{1}{c|}{0.65} & \multicolumn{1}{c|}{0.57} & 0.55 & 0.22 \\ 
                                 & Accuracy  & \multicolumn{1}{c|}{0.43} & \multicolumn{1}{c|}{0.75} & \multicolumn{1}{c|}{0.66} & 0.66 & 0.28 \\ \hline
\multirow{4}{*}{\textbf{SNN}}    & Precision & \multicolumn{1}{c|}{0.67} & \multicolumn{1}{c|}{0.94} & \multicolumn{1}{c|}{0.56} & 0.66 & 0.55 \\ 
                                 & Recall~   & \multicolumn{1}{c|}{0.68} & \multicolumn{1}{c|}{0.98} & \multicolumn{1}{c|}{0.56} & 0.63 & 0.56 \\ 
                                 & F1-Score  & \multicolumn{1}{c|}{0.67} & \multicolumn{1}{c|}{0.95} & \multicolumn{1}{c|}{0.55} & 0.64 & 0.54 \\ 
                                 & Accuracy  & \multicolumn{1}{c|}{0.7}  & \multicolumn{1}{c|}{0.96} & \multicolumn{1}{c|}{0.57} & 0.77 & 0.57 \\ \hline
\multirow{4}{*}{\textbf{MLP}}    & Precision & \multicolumn{1}{c|}{0.7}  & \multicolumn{1}{c|}{0.94} & \multicolumn{1}{c|}{0.6}  & 0.66 & 0.56 \\  
                                 & Recall~   & \multicolumn{1}{c|}{0.63} & \multicolumn{1}{c|}{0.98} & \multicolumn{1}{c|}{0.61} & 0.63 & 0.57 \\  
                                 & F1-Score  & \multicolumn{1}{c|}{0.49} & \multicolumn{1}{c|}{0.95} & \multicolumn{1}{c|}{0.59} & 0.64 & 0.55 \\  
                                 & Accuracy  & \multicolumn{1}{c|}{0.5}  & \multicolumn{1}{c|}{0.96} & \multicolumn{1}{c|}{0.6}  & 0.77 & 0.57 \\ \hline
\multirow{4}{*}{\textbf{1DCNN}}  & Precision & \multicolumn{1}{c|}{0.76} & \multicolumn{1}{c|}{0.94} & \multicolumn{1}{c|}{0.7}  & 0.73 & 0.28 \\  
                                 & Recall~   & \multicolumn{1}{c|}{0.75} & \multicolumn{1}{c|}{0.98} & \multicolumn{1}{c|}{0.69} & 0.58 & 0.43 \\  
                                 & F1-Score  & \multicolumn{1}{c|}{0.7}  & \multicolumn{1}{c|}{0.95} & \multicolumn{1}{c|}{0.63} & 0.58 & 0.32 \\  
                                 & Accuracy  & \multicolumn{1}{c|}{0.7}  & \multicolumn{1}{c|}{0.96} & \multicolumn{1}{c|}{0.63} & 0.79 & 0.43 \\ \hline
\end{tabular}%
}
	\caption{Binary classification using the DL models on benchmark dataset}
		\label{tab:compare-benchmark_DL}
\end{table}
    
    With a high recall (0.98) and accuracy (0.96) in the C1/C3 condition, SNN performs well; but, in the C2/C3 condition, its performance deteriorates, resulting in a mediocre accuracy of 0.57. Although its scores decline for C1/C2, where it attains an accuracy of 0.5, MLP likewise performs best in the C1/C3 scenario, exhibiting high precision (0.94) and accuracy (0.96). 1DCNN shows the highest precision and recall for the C1/C2 condition (0.76 and 0.75, respectively) and achieves a strong accuracy (0.96) in the C1/C3 condition.

	TabNet and SNN are among the models that show moderate to low performance for the C2/C3 condition, indicating that it is difficult to distinguish between these classes. The C1/C3 condition consistently has the highest results across all models, suggesting that it is simpler to differentiate between C1 and C3 than the other circumstances. Both SNN and 1DCNN surpass TabNet in terms of accuracy, peaking at 0.96 in the C1/C3 condition. The maximum accuracy (0.79) for the combined C1, C2, and C3 condition is obtained by 1DCNN, suggesting relative robustness in discriminating the combined classes. In comparison to TabNet and MLP, SNN and 1DCNN perform better overall over a wider range of metrics and situations. Every model exhibits advantages in specific class situations, illustrating the need to adopt a model according to the given classification scenario.

\subsection{Performance Evaluation of the Window-Based Multiclass Classification}\label{window_multiclass}
This section provides performance analysis for the window-based approach on multi-class classification. The provided tables showcase the performance of various machine learning models across different metrics (Precision, Recall, F1, and Accuracy) based on the number of windows (ranging from 5 to 50). Maintaining a single window resulted in data that was too sparse, and therefore not appropriate for deep learning methodology, while varying the window sizes resulted in too much variance between features. Varying window numbers demonstrated the best results. 

\subsubsection{Multi-class classification using ML models}\label{Multi- classification using the ML}
Table \ref{tab_window_ML_MultiClass} presents the performance metrics of four different models—Support Vector Machine (SVM), K-Nearest Neighbors (KNN), Gaussian Naive Bayes (GNB), and Classification and Regression Trees (CART)—across various window sizes (5 to 100). The findings show notable differences in performance between the models and window sizes, with KNN typically performing better, especially at bigger window sizes.

\begin{table*}[!hbtp]
    \centering
    \small
    \begin{tabular}{c|l|l|rrrrrrrrrr} 
    \hline
    \multirow{2}{*}{Model} & \multicolumn{1}{c|}{\multirow{2}{*}{Metric}} & \multirow{2}{*}{\begin{tabular}[c]{@{}l@{}}Baseline\\C1/C2/C3\end{tabular}} & \multicolumn{10}{c}{Number of Windows} \\ 
    \cline{4-13}
     & \multicolumn{1}{c|}{} &  & \multicolumn{1}{r|}{5} & \multicolumn{1}{r|}{10} & \multicolumn{1}{r|}{15} & \multicolumn{1}{r|}{20} & \multicolumn{1}{r|}{30} & \multicolumn{1}{r|}{40} & \multicolumn{1}{r|}{50} & \multicolumn{1}{r|}{70} & \multicolumn{1}{r|}{90} & 100 \\ 
    \hline
    \multirow{4}{*}{SVM}
      & Precision  
        & 0.55 & 0.51 & 0.61 & 0.63 & 0.65 & 0.72 & 0.71 & 0.77 & 0.79 & 0.81 & \textbf{0.85} \\
      & Recall     
        & 0.56 & 0.52 & 0.60 & 0.63 & 0.64 & 0.72 & 0.72 & 0.77 & 0.80 & 0.82 & \textbf{0.86} \\
      & F1-Score          
        & 0.53 & 0.51 & 0.60 & 0.63 & 0.65 & 0.72 & 0.71 & 0.77 & 0.80 & 0.81 & \textbf{0.85} \\
      & Accuracy  
        & 0.58 & 0.51 & 0.60 & 0.64 & 0.65 & 0.72 & 0.72 & 0.77 & 0.80 & 0.81 & \textbf{0.85} \\
    \hline
    \multirow{4}{*}{KNN}
      & Precision 
        & 0.73 & 0.73 & 0.75 & 0.75 & 0.74 & 0.77 & 0.79 & 0.81 & 0.83 & 0.88 & \textbf{0.88} \\
      & Recall    
        & 0.71 & 0.71 & 0.74 & 0.74 & 0.73 & 0.76 & 0.79 & 0.81 & 0.83 & 0.89 & \textbf{0.89} \\
      & F1-Score         
        & 0.72 & 0.71 & 0.74 & 0.74 & 0.73 & 0.76 & 0.79 & 0.81 & 0.83 & 0.88 & \textbf{0.89} \\
      & Accuracy  
        & 0.72 & 0.71 & 0.74 & 0.74 & 0.74 & 0.76 & 0.79 & 0.81 & 0.83 & 0.88 & \textbf{0.89} \\
    \hline
    \multirow{4}{*}{GNB}
      & Precision 
        & 0.76 & \textbf{0.58} & 0.52 & 0.49 & 0.52 & 0.50 & 0.50 & 0.50 & 0.51 & 0.55 & 0.57 \\
      & Recall    
        & 0.62 & \textbf{0.57} & 0.49 & 0.46 & 0.48 & 0.47 & 0.48 & 0.47 & 0.54 & 0.55 & 0.60 \\
      & F1-Score         
        & 0.61 & \textbf{0.56} & 0.46 & 0.43 & 0.46 & 0.43 & 0.45 & 0.43 & 0.51 & 0.50 & 0.53 \\
      & Accuracy  
        & 0.66 & \textbf{0.58} & 0.50 & 0.48 & 0.50 & 0.49 & 0.49 & 0.49 & 0.51 & 0.55 & 0.56 \\
    \hline
    \multirow{4}{*}{CART}
      & Precision  
        & 0.63 & 0.58 & 0.63 & 0.64 & 0.60 & 0.64 & 0.70 & 0.69 & 0.67 & 0.69 & \textbf{0.72} \\
      & Recall    
        & 0.55 & 0.59 & 0.59 & 0.64 & 0.61 & 0.63 & 0.68 & 0.69 & 0.68 & 0.71 & \textbf{0.75} \\
      & F1-Score         
        & 0.56 & 0.58 & 0.59 & 0.64 & 0.60 & 0.63 & 0.69 & 0.69 & 0.67 & 0.69 & \textbf{0.73} \\
      & Accuracy  
        & 0.57 & 0.58 & 0.59 & 0.64 & 0.60 & 0.63 & 0.69 & 0.69 & 0.68 & 0.69 & \textbf{0.74} \\
    \hline
  \end{tabular}
  \caption{Multi-class classification using the ML models on window-based temporal features}
  \label{tab_window_ML_MultiClass}
\end{table*}

Baseline refers to no windows being applied. SVM achieves a precision and recall of 0.85 and 0.86, respectively, for a window size of 100. As the window size grows, SVM gradually improves all metrics. This reflects SVM's efficiency in handling binary classification tasks and shows that it can improve performance with larger datasets. Nevertheless, the beginning values for smaller window sizes (5–10) are comparatively low, indicating an unresponsive start that would prevent it from being used in a particularly demanding environment.
	
KNN, on the other hand, continuously outperforms all other metrics, with precision and recall peaking at 0.88 for window sizes of 90 and 100. This demonstrates KNN's dependability and efficiency in minimizing false positives and false negatives, which makes it a strong option for classification tasks. 
Overall, the findings imply that when using a window-based strategy, KNN and SVM are the best models for binary classification, particularly at higher window sizes. 
    
\subsubsection{Multi-class classification using DL models}\label{Multi-classification using the DL}

 Table \ref{tab_window_DL_MultiClass} provides the performance of multi-class classification using the DL models on temporal features, including MLP, SNN, TabNet, and 1DCNN across various window sizes in binary classification tasks. The results demonstrate variability in model performance, particularly as the window size increases, highlighting the strengths and weaknesses of each model in different scenarios.
	
    \begin{table*}[!h]
    \centering
    \small
    \begin{tabular}{c|l|l|rrrrrrrrrr} 
    \hline
    \multirow{2}{*}{Model} & \multicolumn{1}{c|}{\multirow{2}{*}{Metric}} & \multirow{2}{*}{\begin{tabular}[c]{@{}l@{}}Baseline\\C1/C2/C3\end{tabular}} & \multicolumn{10}{c}{Number of Windows} \\ 
    \cline{4-13}
     & \multicolumn{1}{c|}{} &  & \multicolumn{1}{r|}{5} & \multicolumn{1}{r|}{10} & \multicolumn{1}{r|}{15} & \multicolumn{1}{r|}{20} & \multicolumn{1}{r|}{30} & \multicolumn{1}{r|}{40} & \multicolumn{1}{r|}{50} & \multicolumn{1}{r|}{70} & \multicolumn{1}{r|}{90} & 100 \\ 
    \hline
    \multirow{4}{*}{MLP} & Precision & 0.56 & 0.53 & 0.55 & 0.62 & 0.48 & 0.54 & 0.59 & 0.64 & 0.72 & 0.73 & \textbf{0.78} \\
     & Recall~ & 0.57 & 0.56 & 0.58 & 0.62 & 0.56 & 0.57 & 0.61 & 0.65 & 0.72 & 0.74 & \textbf{0.79} \\
     & F1-Score  & 0.55 & 0.53 & 0.47 & 0.6 & 0.46 & 0.48 & 0.59 & 0.64 & 0.72 & 0.73 & \textbf{0.78} \\
     & Accuracy & 0.57 & 0.55 & 0.57 & 0.61 & 0.53 & 0.55 & 0.61 & 0.66 & 0.72 & 0.74 & \textbf{0.78} \\ 
    \hline
    \multirow{4}{*}{SNN} & Precision & 0.55 & 0.56 & 0.52 & 0.63 & 0.62 & 0.64 & 0.65 & 0.72 & 0.73 & 0.79 & \textbf{0.81} \\
     & Recall~ & 0.56 & 0.58 & 0.55 & 0.63 & 0.62 & 0.64 & 0.66 & 0.72 & 0.74 & 0.79 & \textbf{0.82} \\
     & F1-Score  & 0.54 & 0.56 & 0.52 & 0.63 & 0.62 & 0.64 & 0.65 & 0.72 & 0.74 & 0.79 & \textbf{0.82} \\
     & Accuracy & 0.57 & 0.57 & 0.54 & 0.63 & 0.62 & 0.64 & 0.66 & 0.73 & 0.73 & 0.79 & \textbf{0.81} \\ 
    \hline
    \multirow{4}{*}{TabNet} & Precision & 0.28 & 0.23 & 0.73 & 0.74 & 0.78 & 0.86 & 0.86 & \textbf{0.91} & 0.86 & 0.88 & 0.9 \\
     & Recall~ & 0.36 & 0.37 & 0.72 & 0.74 & 0.78 & 0.86 & 0.85 & \textbf{0.91} & 0.86 & 0.89 & 0.91 \\
     & F1-Score  & 0.22 & 0.26 & 0.72 & 0.74 & 0.78 & 0.86 & 0.85 & \textbf{0.91} & 0.86 & 0.88 & 0.91 \\
     & Accuracy & 0.28 & 0.33 & 0.72 & 0.75 & 0.78 & 0.86 & 0.86 & \textbf{0.91} & 0.85 & 0.88 & 0.91 \\ 
    \hline
    \multirow{4}{*}{1DCNN} & Precision & 0.28 & 0.5 & 0.54 & 0.39 & 0.33 & 0.78 & 0.8 & \textbf{0.84} & 0.13 & 0.15 & 0.3 \\
     & Recall~ & 0.43 & 0.52 & 0.59 & 0.41 & 0.44 & 0.78 & 0.8 & \textbf{0.84} & 0.33 & 0.333 & 0.37 \\
     & F1-Score  & 0.32 & 0.41 & 0.49 & 0.34 & 0.37 & 0.78 & 0.8 & \textbf{0.84} & 0.2 & 0.21 & 0.3 \\
     & Accuracy & 0.43 & 0.49 & 0.58 & 0.43 & 0.48 & 0.78 & 0.8 & \textbf{0.84} & 0.39 & 0.46 & 0.42 \\
    \hline
    \end{tabular}
    \caption{Multi-class classification using the DL models on window-based temporal features}
    \label{tab_window_DL_MultiClass}
    \end{table*}



    


   The MLP performs moderately on all evaluated metrics; as window sizes increase, precision increases between 0.48 and 0.78. Interestingly, at the biggest window size of 100, the model reaches its highest precision (0.78) and recall (0.79), suggesting a better ability to detect true positives with larger datasets. At the same window size, the F1-score peaks at 0.78, indicating a performance that strikes a balance between recall and precision. The performance is less reliable at smaller window sizes (5 to 20), especially for the F1-score, which falls to 0.46 at a window size of 20. This pattern is reflected in the accuracy, which stays below 0.6 for smaller windows but peaks at 0.78 for a window size of 100. In general, the MLP shows a noticeable boost in performance with bigger.
	
	The SNN performs consistently and dependably across a range of criteria, with both precision and recall peaking at 0.82 at a window size of 100. Regardless of window size, the model consistently maintains precision values above 0.50, demonstrating its ability to recognize relevant instances. Additionally, the F1-score steadily improves, reaching a peak of 0.82, demonstrating that the SNN effectively balances recall and precision. Performance shows a slow improvement at lower window sizes (5 to 20), although it is not as strong as at higher sizes. With a maximum accuracy of 0.81 at a window size of 100, the SNN's accuracy is impressive and demonstrates the model's strong generalization with growing data. Overall, the SNN performs well and has good stability, especially at bigger window sizes.
	
	The performance of TabNet is remarkable; it exhibits a noticeable rise in metrics as window sizes increase. With a window size of 5, the precision is low at 0.23, but at a window size of 100, it greatly improves to 0.90. Recall likewise exhibits a positive tendency, rising within the same range from 0.37 to 0.91. This pattern is reflected in the F1-score, which increases from 0.26 to 0.91, suggesting a remarkable trade-off between recall and precision at increasing window sizes. This model's consistent accuracy, which peaks at 0.91 at a window size of 100, indicates that it is very good at processing larger datasets. This implies that, despite its initial difficulties with smaller window numbers, TabNet is well-suited for binary classification tasks, particularly as more data is provided.

	With a noticeable decline in metrics at lower window sizes, the 1DCNN exhibits a diverse performance profile. Recall and precision vary significantly; the maximum precision of 0.84 was achieved at a window size of 50, indicating that it may successfully detect true positives in certain circumstances. However, at higher window sizes (70 to 100), its performance drastically deteriorates, with precision falling to as low as 0.30, indicating that it struggles to find relevant examples in these situations. A similar pattern is shown by the F1-score, which highlights problems with model consistency by peaking at 0.84 and then falling to 0.30 for increasing window sizes. Additionally, the accuracy fluctuates, peaking at 0.84 at window size 50 and falling to a low of 0.39 at window size 70. 
    
    \begin{table}[!hbtp]
    \begin{tabular}{|l|ll|l|}
    \hline
    \multirow{2}{*}{Condition} &
      \multicolumn{2}{c|}{\begin{tabular}[c]{@{}c@{}}Best \\ performers\end{tabular}} &
      \multirow{2}{*}{\begin{tabular}[c]{@{}l@{}}Best performing \\ Approach\end{tabular}} \\ \cline{2-3}
             & \multicolumn{1}{l|}{ML Model} & DL Model &        \\ \hline
    C1/C2    & \multicolumn{1}{l|}{SVM}      & TabNet   & TabNet \\ \hline
    C1/C3 &
      \multicolumn{1}{l|}{KNN} &
      1DCNN &
      \begin{tabular}[c]{@{}l@{}}KNN\\ 1DCNN \\ (tie)\end{tabular} \\ \hline
    C2/C3    & \multicolumn{1}{l|}{KNN}      & TabNet   & TabNet \\ \hline
    C1,C2/C3 & \multicolumn{1}{l|}{KNN}      & TabNet   & TabNet \\ \hline
    \end{tabular}%
        \caption{Comparison of Best Performing ML and DL Models for differing conditions}
        \label{ML vs DL best performers}
    \end{table}

Overall, the 1DCNN appears to perform best at moderate window sizes, while larger windows may lead to decreased efficacy, suggesting a potential issue with overfitting or generalization. Additionally, deep learning models outperform traditional machine learning models across the various window ranges of window sizes, as summarized in Table \ref{ML vs DL best performers}.

In conclusion, there is a large variation in the performance of these models depending on the window numbers employed. The best candidates are SNN and TabNet, which demonstrate good precision, recall, and F1-scores, particularly at bigger window sizes. When compared to SNN and TabNet, MLP shows a slow improvement but still performs moderately. Even though it performs well at specific sizes, the 1DCNN has trouble with wider windows, which suggests that its predictive power is inconsistent. Overall, the task's particular requirements—specifically, the significance of precision vs recall and the type of data being processed—should direct the model selection.

\subsection{Results Discussion}
The results indicate that both machine learning and deep learning models can effectively classify CW using eye-tracking data from the COLET dataset. However, deep learning models outperform traditional machine learning techniques, especially in capturing the temporal and non-linear relationships in the data. The superior performance of the Transformer model suggests that attention mechanisms may be key to improving workload classification in real-time systems. TabNet improves with larger window sizes because it leverages more contextual features through its sequential attention mechanism, enhancing feature selection and pattern recognition, while models like 1D-CNN drop off as their fixed kernel sizes and local receptive fields dilute temporal specificity and introduce noise over extended input ranges.

The comparison of performance metrics across different machine learning models and varying window sizes highlights significant variations in effectiveness for binary classification tasks. 
	
SVM and KNN demonstrated strong and reliable performance throughout various window sizes, with KNN achieving the highest metrics overall. SVM's precision reached 0.85 at a window size of 100, while KNN maintained a high level of performance with precision and recall both peaking at 0.89, indicating its robustness in this classification task. The KNN model’s consistency across metrics illustrates its effectiveness, particularly in maintaining a balanced trade-off between precision and recall, making it a strong contender for binary classification tasks. On the contrary, models like GNB and 1DCNN struggled to deliver competitive results, especially at smaller window sizes. GNB showed the lowest overall performance, with its highest accuracy only reaching 0.58, while 1DCNN exhibited erratic performance, particularly at larger window sizes, where accuracy dropped significantly to 0.30. 
    
In contrast, TabNet displayed an interesting trend with a very low precision of 0.23 at a window size of 5, but it subsequently improved dramatically, achieving a maximum accuracy of 0.91 at a window size of 100. This suggests that TabNet may require a larger amount of data to perform effectively, aligning with its characteristic strengths in complex data patterns. Overall, the results indicate that while some models excel across a range of parameters, others are highly sensitive to the choice of window size and require careful selection based on the specific application needs. The models performed better at a greater number of windows and at more granular temporal features. The impact of temporal features on model performance (both machine learning and deep learning) highlights the importance of examining eye movement features at a granular level. Additionally, this highlights that aggregate features are insufficient for detecting CW, and future research should further explore this approach. 
	
\section{Conclusion}\label{sec_conclusion}
This research presents a window-based approach to CW classification using eye-tracking data from the COLET dataset. By applying ML and DL models, we demonstrate that eye-tracking metrics can serve as reliable indicators of CW. DL models show promising performance in accurately classifying workload, paving the way for real-time cognitive state monitoring in critical applications. Additionally, the window-based approach allows for efficient feature generation from time-series data, ensuring that important temporal patterns are preserved. Future work can explore more advanced techniques for window selection, including adaptive window sizes based on workload transitions. 
	
    \bibliographystyle{elsarticle-num} 
    \bibliography{example}  
	
\end{document}